\relax
\documentclass[letterpaper]{article} 
\usepackage{iclr2022_conference,times}
\usepackage{times}  
\usepackage{helvet}  
\usepackage{courier}  
\usepackage[hyphens]{url} 
\usepackage{graphicx} 
\usepackage{natbib}  
\usepackage{caption} 
\DeclareCaptionStyle{ruled}{labelfont=normalfont,labelsep=colon,strut=off} %
\usepackage{algorithm}
\usepackage{algorithmic}

\usepackage{newfloat}
\usepackage{listings}
\floatstyle{ruled}
\newfloat{listing}{tb}{lst}{}
\floatname{listing}{Listing}

\usepackage{amsmath,amssymb}
\usepackage[capitalise]{cleveref}

\newcommand{\GH}{{Github}\xspace}
\newcommand\p{\ensuremath{{\mathrm{F}_u}}}
\newcommand\pref{\ensuremath{{\mathrm{F}_\mathit{ref}}}\xspace}
\newcommand\popt{\ensuremath{{\mathrm{F}_{o}}}\xspace}
\newcommand\s{\ensuremath{{\mathrm{S}}}\xspace}
\newcommand\popthat{\ensuremath{\hat{\mathrm{F}}_{o}}}

\newcommand\IO{\ensuremath{\{ IO \}_{k=1}^K}\xspace}

\newcommand\C{\ensuremath{\mathcal{C}}\xspace}
\newcommand\V{\ensuremath{\mathcal{V}}\xspace}

\usepackage{pifont}

\usepackage{subcaption}
\usepackage{booktabs}
\usepackage{multirow}
\usepackage{xspace}
\usepackage{xcolor, soul}
\usepackage{multibib}

\newcommand{\gcc}{\texttt{gcc}\xspace}
\newcommand{\ozero}{\texttt{-O0}\xspace}
\newcommand{\othree}{\texttt{-O3}\xspace}

\newcommand{\register}[1]{\lstinline{\%#1}}

\newcommand{\rax}{\register{rax}\xspace}
\newcommand{\cltq}{\lstinline{cltq}\xspace}
\newcommand{\movslq}{\lstinline{movslq}\xspace}

\lstdefinelanguage
  [x64]{Assembler}
  [x86masm]{Assembler}
  {basicstyle=\ttfamily\bfseries\scriptsize,
  frame=single,
  keepspaces=true,
  framesep=4pt,
  numbers=none,
  morekeywords={addl,popq,pushq,movq,movl,retn,rbp,rsp},
  xleftmargin=3pt, 
  xrightmargin=3pt, 
  tabsize=1}
\lstdefinelanguage
    [hexrays]{Assembler}
    [x86masm]{Assembler}
    {morekeywords={addl,popq,pushq,movq,movl,retn,rbp,rsp},deletendkeywords={dword,ptr,short},morendkeywords={var1,var2},
    }

\title{Learning to Superoptimize Real-world Programs}

\author{Alex Shypula \\
MIT CSAIL \thanks{ Work completed while at Carnegie Mellon University} \\
\texttt{shypula@mit.edu} \\
\And
Pengcheng Yin \\
Google Research *\\
\texttt{pcyin@google.com} \\
\And
Jeremy Lacomis, Claire Le Goues, Edward Schwartz, and Graham Neubig\\
Carnegie Mellon University \\
\texttt{jlacomis@cs.cmu.edu, clegoues@cs.cmu.edu} \\ \texttt{eschwartz@cert.org, gneubig@cs.cmu.edu}
}

\usepackage{bibentry}
\iclrfinalcopy 

\begin{document}

\maketitle 

\begin{abstract}

  Program optimization is the process of modifying software to execute
  more efficiently.
  %
  %
  \emph{Superoptimizers} attempt to find the optimal
  program by employing significantly more expensive search and
  constraint solving techniques.
  Generally, these methods do not scale well to programs in real
  development scenarios, and as a result superoptimization has largely
  been confined to small-scale, domain-specific, and/or synthetic
  program benchmarks.
  In this paper, we propose a framework to learn to superoptimize
  real-world programs by using neural sequence-to-sequence models.
  We created a dataset consisting of over 25K
  real-world x86-64 assembly functions mined from open-source projects and propose an approach, \textbf{S}elf
  \textbf{I}mitation \textbf{L}earning for \textbf{O}ptimization (\textbf{SILO}) %
  that is easy to implement and outperforms a standard policy gradient
  learning approach on our dataset. Our method, SILO, superoptimizes 5.9\% of our test set when compared with the \gcc version 10.3 compiler's 
  aggressive optimization level %
  \othree. We also report that SILO's rate of superoptimization on our test set is over five times that of a standard policy gradient approach and a model pre-trained on compiler optimization demonstration.%
\end{abstract}

\noindent 
\section{Introduction}

Program optimization is a classical problem in computer science that has existed for over 50 years \cite{mckeeman1965peephole, allen1971}.
The standard tool for generating efficient programs is an \emph{optimizing compiler} that not only converts human-written programs into executable machine code, but also performs a number of semantics-preserving code transformations to increase speed, reduce energy consumption, or improve memory footprint \citep{dragonbook}. Most optimizing compilers use semantics-preserving heuristic-based optimizations. These optimizing transformations generally need to be written by experts for an individual compiler, and are applied to an intermediate representation of the code produced over the course of transforming high-level code into executable machine code. 
In an effort to automatically create optimized programs that surpass human-defined heuristics, the research community has pioneered automated optimization methods, or ``superoptimizers.'' 
These superoptimizers may outperform compiler-based optimizations, but are difficult to employ in practice. 
Research in machine learning-based program optimization
remains relatively under-explored, especially in light of the progress deep learning models have made in reasoning about and generating code for a variety
of tasks~\cite{allamanis2018survey, chen2021evaluating}. 

%

In this work, we investigate the ability of deep neural networks to
optimize real-world programs mined from \GH. 
For this, we created \textbf{Big Assembly}, a dataset consisting of over 25K functions in x86-64 assembly mined from online open-source projects from \GH, which enables experimentation on large-scale optimization of real-world programs. 
We also propose an easy to implement algorithm \textbf{S}elf
  \textbf{I}mitation \textbf{L}earning for \textbf{O}ptimization (\textbf{SILO}) that progressively improves its superoptimization ability with training. Our results indicate that it superoptimizes 5.9\% of our test set beyond \gcc \othree, over five times the rate of a model pre-trained on the outputs of an optimizing compiler as well as a model fine-tuned with policy gradient methods. Instead of focusing on customized search method unique to a language's implementation and semantics, our methodology relies on a dataset of demonstrations for pre-training as well as a test case generator, a sandbox for executing programs, and a method for verifying program equivalence. 
  
\section{Problem Formulation}

The superoptimization task is, when given a specification for a program \s and a reference program \pref (which meets the specification),
to generate an optimized program \popt that runs more efficiently and
is equivalent to the reference program \pref.
In this paper we focus on program optimization at the assembly level,
so \s, \pref, and \popt are all programs written in X86-64 assembly code. \s is a program with no optimizations applied at all and its purpose is to demonstrate desired program semantics. \pref is an assembly program produced by the optimizing compiler \gcc at its aggressive \othree optimization level. 

%
%
Specifically, our goal is to learn a model
$f_{\theta}: \s \mapsto \popt$ such that a model-generated (optimized)
program $\hat{\mathrm{F}}_o$ attains lower cost, and ideally minimal cost, under a cost function
$\mathcal{C}(\cdot)$ evaluated on a suite of $K$ input-output test
cases $\{ IO \}_{k=1}^K$ (for example energy consumption or runtime). Here, $I$ represents the hardware state prior to executing the
program (i.e., input) and $O$ represents the hardware state after
executing the program (i.e., output). This model-generated program must meet the specification, which is determined by a verification function $\mathcal{V}(\cdot) \in \{ 0, 1 \}$. More details about \C and \V are located in \cref{sec:evaluation}. The learned model's objective is then to produce rewrites that meet the condition: 

\begin{equation}
    \label{eqn:optimizaiton_goal}
    \begin{split} 
        \mathcal{C} \Big(\popthat; \{IO_k\}_{k=1}^k \Big)  \ 
        < \ 
        \mathcal{C} \Big(\pref; \{IO_k\}_{k=1}^K \Big) \;\;
         s.t. \;\; \mathcal{V}( \ 
                        \popthat, \s ) \ 
                        = 0  
    \end{split}
\end{equation}

In order to train our model on some of the optimizations that are present
in modern compilers in a supervised manner and to improve it by learning from experience a dataset is necessary.  Our training set $D_o$ therefore consists of $N$ tuples of (1) an I/O
test suite $\{IO_k\}_{k=1}^K$, (2) a compiled and unoptimized program specification \s, (3) and a compiled and aggressively
optimized program \pref: 

\begin{equation} 
    \label{eqn:init_dataset}
        D_o = \
                \bigg\{
                    \Big( \
                        \{IO_k^i\}_{k=1...K}, \
                        \s^i, \
                        \mathrm{F}^i_{\mathit{ref}}, \
                    \Big) \
                \bigg\}_{i = 1...N} \\
\end{equation}

In \cref{sec:learning} we explain our methodology for learning to superoptimize programs; before this, however, we first introduce our dataset for optimizing real-world programs in \cref{sec:benchmark}.

\section{Big Assembly, a Dataset Mined from from Real World Code}
\label{sec:benchmark}

There is no standard benchmark for evaluating superoptimization research.
Some researchers have evaluated on randomly-generated programs in a
simplified domain specific language \cite{chen2019learning, shi2020},
while others have tested on small and hand-picked
programs~\cite{joshi2002denali,gulwani2011synthesis,
  churchill2017sound}.
While these datasets are sufficient for demonstrating methodological
capabilities, they do not necessarily reflect the properties of real
code, and thus do not predict their performance relative to modern
optimizing compilers on real-world programs. Lastly, small-scale
benchmarks are insufficient for data-hungry modern deep learning.

We created a dataset, which we will refer to as \textbf{Big Assembly}, consisting of of 25,141 functions in x86-64 assembly code collected by using \gcc to compile programs both with (\othree) and without (\ozero) aggressive optimizations.
We started by collecting 1.61 million functions from open source projects on \GH that were written in C.
Of these 1.61M, we were able to mine testcases for a dataset of over 100,000 functions. We performed two stages of sanity checks and analysis to compute a conservative approximation of the live out registers.  The first involved using SMT solvers to find equivalence between the \gcc \ozero function used as the specification \s and the \gcc \othree function used as \pref and filtered out trivial programs such as those equivaent to \texttt{return 0}, this reduced our dataset to 77,813 functions. We then performed a similar pass, but instead using a test case suite of randomly generated inputs and their corresponding outputs \IO reducing our total dataset to 25,141 functions (19,819 train, 2,193 dev, 3129 test). 
%
For verification, test case generation, and program instrumentation, we used artifacts from the \textsc{STOKE}\footnote{\url{https://github.com/StanfordPL/stoke}} project with
additional modifications. 

\cref{tab:superopt_benchmarks} compares the basic properties of our
dataset to those from existing work; beyond its size, our dataset it notable because it consists entirely of functions mined from real world
codebases.  It also contains examples with more complex operations such as
SIMD instructions, branching, and loops.
Additional details on
how the dataset was collected are available in the supplementary
materials section.

\begin{table}[t]
\caption{A list of program optimization benchmarks from machine learning and programming languages / systems works. The three criteria for evaluating listed are the number of individual examples in the benchmark (Sz.), are the programs written by humans (H.), are the programs found ``in the wild" (i.e. in open source projects) (R.W.), and does the benchmark contain either branching or control flow (CTL.) }
\label{tab:superopt_benchmarks}
\vskip 0.15in
\begin{center}
\begin{small}
\begin{sc}
\begin{tabular}{lcccc}
\toprule
Dataset & Sz & H. & R.W. & CTL. \\
\midrule
\citet{shi2020}  & 12,000 & \ding{55} & \ding{55} & \ding{55}\\
\citet{gulwani2011synthesis} & 25 & \ding{51} & \ding{55} & \ding{55} \\
\citet{churchill2017sound}  & 13 & \ding{51} & \ding{51} & \ding{51} \\
Ours  & 25,141 & \ding{51} & \ding{51} & \ding{51} \\
\bottomrule
\end{tabular}
\end{sc}
\end{small}
\end{center}
\vskip -0.1in
\end{table}

\section{Learning Program Optimizations}
\label{sec:learning}


\subsection{Neural Program Optimizer}

Our program optimization model $f_{\theta}$ is a neural sequence-to-sequence network, where the input specification (unoptimized program) and output (optimized program) are represented as sequences of tokens.
Specifically, $f_{\theta}$ is parameterized with a standard Transformer-based encoder-decoder model \citep{vaswani2017attention}.

\subsection{Learning Algorithms}
\label{sec:hillclimbing}

For learning to optimize, we develop a two-stage learning approach.
First, in a \emph{pre-training} stage, to capture commonly-used optimization heuristics adopted by existing optimizing compilers, we use supervised learning to train the model on the mined corpus $D_o$ of \gcc-optimized programs (E.q.~\ref{eqn:init_dataset}) described in \cref{sec:benchmark}. 
Next, to discover more efficient optimization strategies, we investigate fine-tuning using policy-gradient methods and a propose an iterative learning approach, SILO. 

\paragraph{Policy Gradient Approach}

As in \cref{eqn:optimizaiton_goal}, our goal is to synthesize a correct program $\popthat$ 
verified by $\mathcal{V}$ 
that outperforms a reference program $\pref$ on the cost function \C. 
An intuitive choice is to use policy gradient methods to learn a policy that directly minimizes our cost function \C and produces correct programs under \V in expectation. 
Specifically, we express this dual objective via the Lagrangian relaxation:
\begin{equation}
    \label{eqn:rl_goal}
    \begin{split}
        \mathbf{J}(\popthat) = \mathcal{C} \Big(\popthat; \{IO_k\}_{k=1}^k \Big) + \lambda \; \mathcal{V}( \ 
                        \popthat, \s )  
    \end{split}.
\end{equation}

A commonly used policy gradient approach is REINFORCE with baseline \cite{williams1992simple}. Based on our minimization objective, we can express the loss using the following equation, where $b(\s)$ is a baseline value for the given specification and $p(a_t| a_{<t}; \s )$ is the model-given probability for generating a token at time step $t$ for sequence $\popthat$. In a traditional reinforcement learning context one might seek to perform gradient ascent on the following term; however, because we are trying to minimize the objective function, we perform gradient descent instead.

\begin{equation}
    \label{eqn:rl_loss}
    \begin{split}
        \mathcal{L} = \sum_{t=1}^T \log p(a_t | a_{<t}; \s ) \: \big( \mathbf{J}(\popthat) - b(\s) \big)
    \end{split}
\end{equation}

\paragraph{\textbf{S}elf \textbf{I}mitation \textbf{L}earning for \textbf{O}ptimization (SILO)} 

\cref{alg:hill_climbing} illustrates our SILO learning approach.
It consists of two steps, an exploration step (lines 4-11) where the model seeks to discover alternative optimizations that are more efficient than the compiler generated targets used in pre-training, and a learning step (lines 12-13), where the model parameters are updated using newly discovered optimized programs.
First, in the exploration step, an exploration batch $B_{\mathrm{ex}}$ is sampled from the dataset $D$ initialized with program specifications (in our case, unoptimized assembly programs) $\mathrm\s^i$ and their compiler-optimized outputs $\mathrm{F}^i_{\mathit{ref}}$. 
For each input specification $\mathrm\s^i$ in $B_{\mathrm{ex}}$, we sample a model-predicted optimization $\hat{\mathrm{F}}^i_o$, and execute $\hat{\mathrm{F}}^i_o$ on the I/O test suite \IO to compute the cost function \C.
If any of the new samples are both functionally equivalent by our verification function \V and also achieve a lower cost under  %
\C compared to the compiler-optimized targets in the original dataset, the compiler-optimized target in the dataset is then replaced with the model's newly-discovered optimal rewrite. 
After the exploration step is taken, in the learning step, a separate training batch $B_{\mathrm{tr}}$ is sampled for maximum-likelihood training from the dataset which may now contain model-optimized targets. 

Self-imitation learning \cite{oh2018self} is an off-policy reinforcement learning algorithm intended to help agents solve challenging exploration problems by learning from good past actions.
Intuitively, besides the ordinary on-policy reinforcement learning using the latest model-predicted actions, the model is also trained on historical states $s$ and actions $a$ that achieve high rewards $R$ using a cross-entropy loss:
\begin{equation}
    \label{eqn:sil_loss}
    \begin{split}
        \mathcal{L}_{\text {sil }} &=-\log p(a | s) \; \max \big( (R-V_{\theta}(s)), 0 \big)
    \end{split}
\end{equation}
where each sample $\langle a, s \rangle$ is weighted by how much better the off-policy return was compared to the learned baseline $V_{\theta}(s)$. 
Our algorithm, SILO, has a few differences with standard self-imitation learning. First, for sequences that outperform $\pref$ we omit a learned value function and train on the entire sequence using cross-entropy loss, as opposed to individual actions. Additionally, we do not interpolate our loss with an on-policy reinforcement learning algorithm. Rather, we avoid the policy gradient altogether, and instead train only on the best sequence found so far in our dataset, be it the compiler-optimized outputs $\pref$ or a sequence discovered that outperforms it $\popt$.
This is also broadly related to the ``hard'' EM algorithm that uses the currently best model-predicted results as optimization targets \cite{kearns1998information}.

\begin{algorithm}
\begin{algorithmic}[1]
\STATE Initialize model $f$ parameters $\theta$ from pre-trained model \\
\STATE Initialize dataset of program function pairs and test cases: \\  
    $D = D_o =  \Big\{ \big( \
    \{IO^i\}_{k=1}^K, \
    \textrm\s^i, \
    \textrm{F}_{\mathit{ref}}^i, \
    \big) \ 
    \Big\}_{i=1...N}$ \\
 \WHILE{budget not exhausted}
  \STATE Sample a batch $B_{\mathrm{ex}}$ from $D_o$\\
  \FOR{ $\big( \
    \{IO^i\}_{k=1}^K, \
    \textrm\s^i, \
    \textrm{F}_{\mathit{ref}}^i, \
    \big)$  in $B_{\mathrm{ex}}$} 
    \STATE  sample $\hat{\textrm{F}}_{o}^i \; \sim \; f_{\theta}(\s^i)$ \\
    \STATE calculate 
    $\mathcal{C}(\hat{\textrm{F}}_{o}^i)$, 
    and $\mathcal{V} (\hat{\textrm{F}}_{o}^i, \textrm\s^i)$ 
    \IF{$\mathcal{V} (\hat{\textrm{F}}_{o}^i, \textrm\s^i) = 0$ and \ 
    $\mathcal{C}(\hat{\textrm{F}}_{o}^i) < \ 
    \mathcal{C}(\textrm{F}_{\mathit{ref}}^i) $ }
        \STATE  Replace $\textrm{F}_{\mathit{ref}}^i$ with sample $\hat{\textrm{F}}_{o}^i$ in $D$ \\
    \ENDIF
  \ENDFOR
  \STATE Sample a batch $B_{\mathrm{tr}}$ from $D$\\
  \STATE Update $\theta$ via supervised learning on $B_{\mathrm{tr}}$ from $D$ \\
 \ENDWHILE
 \caption{SILO for Program Optimization}
 \label{alg:hill_climbing}
\end{algorithmic}
\end{algorithm}

\subsection{Actor-Learner Architecture, Model Configuration, and Training}
One hurdle to performing program optimization at scale is that the the time required to evaluate $\mathcal{C}$ and $\mathcal{V}$ is costly, limiting the model's throughput of learning examples.
To alleviate this bottleneck, we utilize an actor-learner set up  \citep{liang2018memory, espeholt2018impala}. Additional details on our actor-learner set up are provided in the appendix. 

%

Our neural superoptimizer $f_{\theta}$ uses a 3-layer transformer encoder-decoder with embedding dimension of 512 with 8 attention-heads. We utilize the Adam optimizer \cite{kingma2014adam} with the inverse square root schedule from \citet{vaswani2017attention}. We pre-trained the model for 88K steps and subsequently performed our fine-tuning algorithms for an additional 5K steps. We additionally use \texttt{SentencePiece}\footnote{ \url{https://github.com/google/sentencepiece}} to pre-process the assembly with byte-pair encoding \citep{sennrich2015neural}. Additional details on training hyperparameters and data preprocessing are located in our supplementary materials section.

\section{Evaluation}
\label{sec:evaluation}

In our experiments, we test our methods by taking our final model, and generate 10 model-optimized candidates through beam search, and then calculate \C and \V for each. We then report results based on the best result of all 10 candidate programs.

A primary concern in constructing the verification function \V in \cref{eqn:optimizaiton_goal} is the undecidability of program equivalence for programs with control flow such as loops. If using testcases for equivalence \V as well as the cost function \C challenges also lie in utilizing a testcase suite \IO for either benchmarking a program's performance or coming up with an approximate estimate for performance. 

\subsection{Measuring Program Correctness}

A key claim of this work is that our superoptimizer outputs correct programs --- that is, programs that are more efficient, but still semantically equivalent, to the input programs.  Program equivalence is undecidable in the general case, motivating a complementary set of mechanisms for verifying output program correctness.
In our experiments, we confirm output program correctness for our verification function \V in two ways.  First, we run synthesized programs on the provided test cases.  Second, we formally verify correctness using two solver based verifiers: the bounded SMT solver based verifier from the standard \textsc{STOKE} project artifacts, and an additional verifier available from the artifacts in the program verification work in \citet{churchill2019semantic}. 
These program verifiers are based on the state-of-the-art Z3 SMT solver \cite{de2008z3}; SMT (``Satisfiability Modulo Theories") solvers decide the satisfiability of boolean formulas with respect to a set of underlying theories (such as the theory of integer arithmetic). 

We use both test cases and the two verifiers for several reasons.  High coverage test suites are informative in terms of program correctness, but intrinsically incomplete.  Meanwhile, verifiers do not always scale, especially to programs with arbitrary numbers of loop iterations to ensure termination.
As is standard we configured a maximum bound of $b$ (set to 4) loop iterations, and the process does not timeout by taking over $T$ seconds (set to 150). 
Verification is thus also incomplete past those bounds, and limited by the correctness of the verifiers' underlying models.  

Indeed, we manually observed cases where our fine-tuning methods exploited either gaps in test suite coverage, or bugs in the verifiers' models of x86-64 semantics.\footnote{We have since reached out to the authors of \cite{churchill2019semantic} regarding issues we have found}
Motivated by this, we use both testcases and the two verifiers for additional robustness. %
While this would intuitively help mitigate spuriousness during evaluation, it could still remain an issue when evaluating optimization of open-domain programs.
As we later explain in \cref{sec:BigAssemblyResults}, we also resort to human verification to get reliable results when reporting model performance on test sets.

\subsection{Measuring Program Performance} 
We follow previous work on superoptimization of x86-64 assembly and primarily calculate the cost function $\mathcal{C}$ (E.q.~\ref{eqn:optimizaiton_goal}) as a static heuristic approximation of expected CPU-clock cycles. 
We compute the sum of both performance cost functions from \citet{schkufza2013stochastic} and \citet{churchill2017sound}. 
The former is a sum of all expected latencies for all instructions in the assembly function (denoted as $\mathcal{C}_\textrm{all}$), while the latter computes expected latencies only using executed instructions ($\mathcal{C}_\textrm{exe}$) from the randomly generated test suite \IO.
$\mathcal{C}_\textrm{exe}$ is a better approximation, especially for functions that contain loop constructs, while $\mathcal{C}_\textrm{all}$ may additionally penalize redundant instructions that are not executed. 
Expected latencies were calculated by the authors of STOKE for the Intel Haswell architecture by benchmarking and measuring instructions for servers. 


\section{Application to Hacker's Delight}

We apply our methods to the 25 functions chosen from the \textsc{Hacker's Delight} benchmark \cite{warren2002hacker}, first used in \citet{gulwani2011synthesis} for program synthesis and later in \citet{schkufza2013stochastic} for x86-64 program superoptimization. In the latter work, authors express they were able to either match or outperform \gcc \othree when provided programs compiled with \texttt{LLVM -O0}. The superoptimization benchmark consists of bit-vector manipulation challenges such as ``take the absolute value of x." Before evaluating on our large scale Big Assembly dataset in \cref{sec:BigAssemblyResults}, we perform controlled experiments on \textsc{Hacker's Delight} allowing for interpretable optimizations and consistency with prior work. 

\newcommand{\othreecodecaption}{-O3 code\xspace}
\newcommand{\modelcodecaption}{Model-optimized code\xspace}
\newcommand{\codeboxwidth}{.35\textwidth}
\newcommand{\pagewidth}{.55\textwidth}
\sethlcolor{pink} 

\lstset{escapeinside={(*@}{@*)}}



\begin{figure}[h]
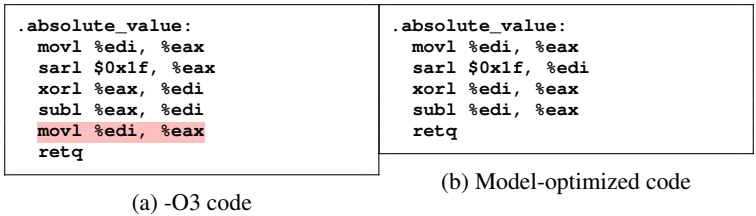

    \begin{subfigure}[t] {\codeboxwidth}
    \begin{lstlisting}[language={[x64]Assembler}]
.absolute_value:
  movl %edi, %eax
  sarl $0x1f, %eax
  xorl %eax, %edi
  subl %eax, %edi
  (*@\hl{movl \%edi, \%eax}@*)
  retq
\end{lstlisting}
    \caption{\othreecodecaption}
    \end{subfigure}
    \begin{subfigure}[t]{\codeboxwidth}
    \begin{lstlisting}
.absolute_value:
  movl %edi, %eax
  sarl $0x1f, %edi
  xorl %edi, %eax
  subl %edi, %eax
  retq
\end{lstlisting}
    \caption{\modelcodecaption} 
    \end{subfigure}
    \centering
    \caption{An example of the absolute value function from \textsc{Hacker's Delight} optimized by \gcc \othree on the left and the trained models on the right. For the right example, the same output was witnessed in the results of both fine-tuning experiments. The model-optimized code demonstrates superior register allocation.}
    \label{fig:hacker_abs_val}

\end{figure}

\paragraph{Results}

We examined the results of REINFORCE and SILO with respect to two quantities: (1) the number of programs where a superoptimized version was found at least once during the training process, and (2) the number of programs for which a superoptimized version was found within the top-10 hypotheses generated by beam search from the last model at the end of training.
Regarding the former metric, REINFORCE and SILO respectively found 3 and 2 superoptimized programs during the training. 
For the latter metric, the final models produced by REINFORCE and SILO output 1 and 2 superoptimized programs respectively.
These results indicate that while the policy gradient methods may have a wider breadth for exploration during the training process compared to SILO, policy gradient methods may also be less stable in their final solutions.

\section{Application to Big Assembly}
\label{sec:BigAssemblyResults}

In applying our methods to the Big Assembly dataset, we followed the same general experimental setup as \textsc{Hacker's Delight}; however, we evaluated our results on held-out sets. As mentioned in \cref{sec:evaluation}, we observed cases where our models exploited bugs in the verifiers' models of x86-64 semantics, thus we also incorporated manual human evaluation into our reporting methodology for the dataset. 

For both learning methods, we chose the best model of the ones checkpointed every 1K steps during fine-tuning. We did this by choosing the model with the highest proportion of programs that were superoptimized on a randomly-sampled subset of 329 functions from the validation set according to our cost function \C, correct according to our verification methodology \V, and correct again by manual human evaluation. 

Using the chosen model, we then evaluated performance on the test set by manually checking all reported superoptimizations for correctness \footnote{Two authors manually reviewed over 80\% of all reported superoptimizations across all 10 beams and checked such exampled for agreement. The additional ~20\% was manually checked by only one author. The authors checked only for false-positives; therefore, no outputs that were deemed incorrect to, equal-to or suboptimal to \othree automatically were not reviewed.}. In \cref{tab:bigass_results} we report (1) the proportion of the entire test set that was superoptimzied according to our automatic methods discussed in \cref{sec:evaluation} as well as (2) the actual proportion that we manually verified to be correct. 

\begin{figure}
    \centering
    \includegraphics[scale=.40]{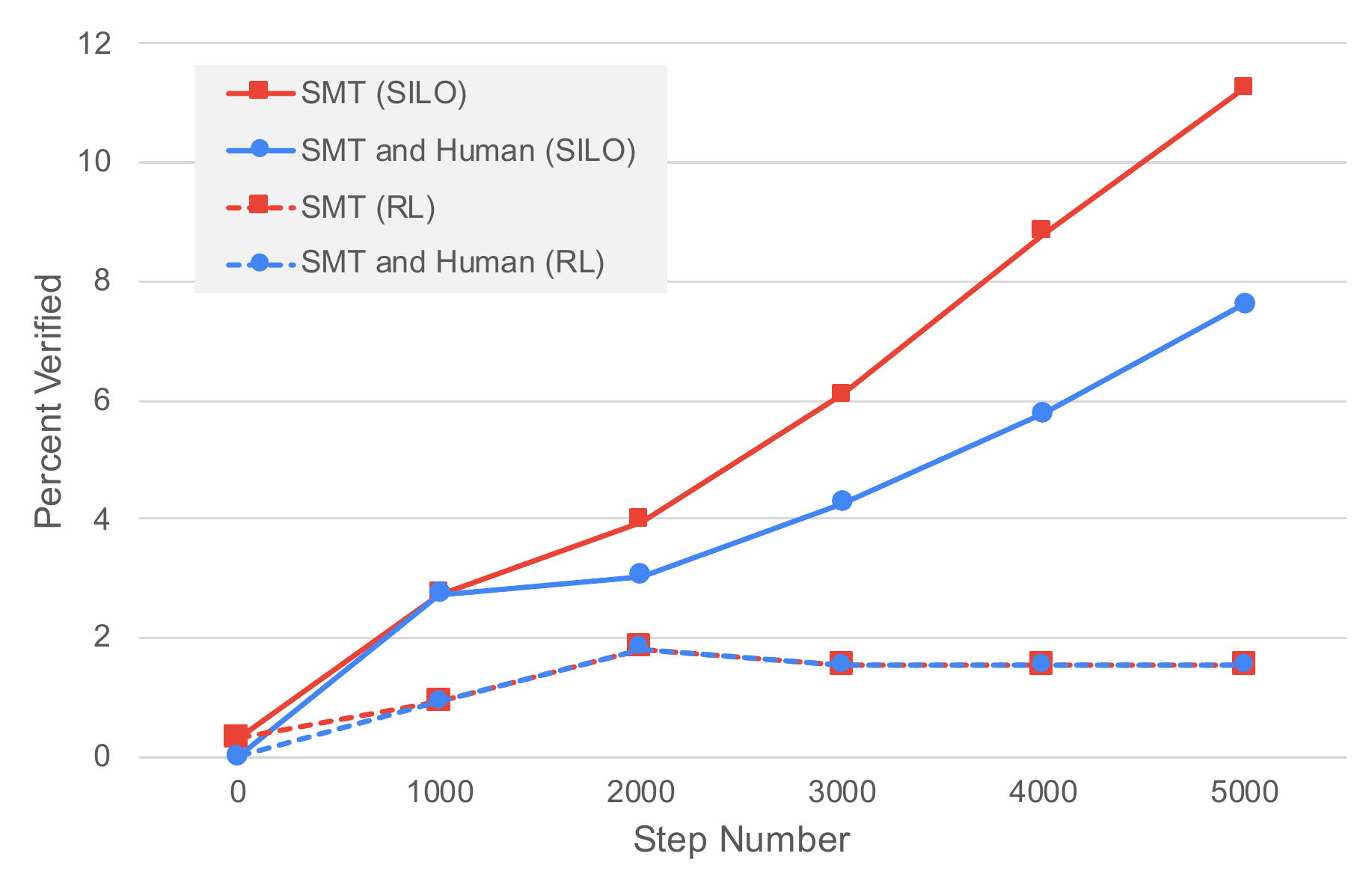}
    \caption{A plot reflecting the proportion of the validation set sub-sample population (329 programs) superoptimized every 1000 steps of training.
    }
    \label{fig:validation_perf}
\end{figure}

\paragraph{Results} We witness that our proposed algorithm SILO far outperformed REINFORCE with baseline on this task: on the test set, SILO superoptimized 5.9\% of programs and REINFORCE with baseline superoptimized 0.9\%.  We also note that despite the fact we used two separate SMT based verifiers to prove correctness, our SILO approach was capable of finding and learning generalizable exploits in a manner that REINFORCE did not. 

In our study of manually verifying assembly programs, we witnessed that across all earlier and later stages of training, the REINFORCE model consistently superoptimized the same 5 to 6 programs in the validation set; in other words, it did not seem to apply superoptimization patterns to any new programs or learn any new superoptimization patterns. In contrast, the SILO model consistently increased the number of programs it superoptimized in the held-out set over time: it seemed to broaden its capacity to apply patterns to new programs and simultaneously learn different strategies to superoptimize and even exploit the verifier; this trend is reflected in the \cref{fig:validation_perf} plot. %

\begin{table}[h]
\small
\centering
\caption{Test set results on the Big Assembly dataset comparing the pre-trained model, SILO and REINFORCE with baseline. The first column (SMT Ver.) reports the proportion of programs that beat the \gcc \othree baseline and verify using our automated evaluation methods, while the second column (SMT + Human Ver.) reports our expected proportion of programs that additionally pass a human evaluation step.}
\vskip 0.15in
\begin{tabular}{lrrrr}
\toprule
\textbf{Model} & \multicolumn{1}{c}{\textbf{SMT Ver.}}  & \multicolumn{1}{c}{\textbf{SMT + Human Ver.}} \\
\midrule
\textsc{Pre-train} &  1.2\%  & 1.0\%\\
\textsc{SILO}        & 8.3\%   & 5.9\%\\
\textsc{REINFORCE}        &  0.9\%  & 0.9\%\\
\bottomrule
\end{tabular}
\label{tab:bigass_results}
\end{table}

We hypothesize the large difference in performance can be attributed to a reward space that is both very sparse and noisy. It is sparse, because program superoptimziations are hard to find: while training, we saw that less than 1 in every 1,000 samples the REINFORCE model made were program superoptimizations. %
The reward space in this task is noisy, because a minuscule change in the output text can have an extreme impact on program semantics and syntactic correctness. %
We believe, without a method to re-learn from past experience, the on-policy REINFORCE algorithm struggles to find the signal in the noise.

\begin{figure}

    \begin{subfigure}[t]{\codeboxwidth}
    \begin{lstlisting}[language={[x64]Assembler}]
.popEntry.s:
  (*@\hl{movl 0x4(\%rdi), \%eax}@*)
  (*@\hl{subl \$0x1, \%eax}@*)
  (*@\hl{movl \%eax, 0x4(\%rdi)}@*)
  cltq 
  leaq (%rax,%rax,2), %rdx
  movq 0x8(%rdi), %rax
  leaq (%rax,%rdx,4), %rax
  retq
\end{lstlisting}
    \caption{\othreecodecaption}
    \end{subfigure}
    \begin{subfigure}[t]{\codeboxwidth}
    \begin{lstlisting}
.popEntry.s:
  (*@\hl{subl \$0x1, 0x4(\%rdi)}@*)
  movq 0x8(%rdi), %rax
  movslq 0x4(%rdi), %rdx
  leaq (%rdx,%rdx,2), %rdx
  leaq (%rax,%rdx,4), %rax
  retq
\end{lstlisting}
    \caption{\modelcodecaption} 
    \end{subfigure}
    \centering
    \caption{An example of a program from the Big Assembly dataset superoptimized with the SILO-trained model. Here a subtraction is performed in memory to eliminate the instructions performing the subtraction in \rax (in red) and storing it back in memory. This approach is followed by \movslq instead of \cltq along with modified register allocation to accommodate the changes.}
    \label{fig:bigass_example}

\end{figure}

\section{Related Work}

\paragraph{Program Optimization}

The general undecidability of program equivalence means that there may always be room for improvement in optimizing programs~\citep{rice1953}. This is especially true as hardware options and performance goals become more diverse: what transformations are best for a scenario may vary greatly on performance objectives such as such as energy consumption or runtime or other factors. 

%

State-of-the-art methods for superoptimization either rely on search-based procedures \citep{schkufza2013stochastic}, or constraint-based methods \citep{sasnauskas2017}. However, these methods have difficulty scaling to larger problems, and as a result, typically do not meet the performance requirements of real development scenarios at compile time.

\paragraph{Machine-Learning-Based Program Optimization} 

Perhaps the closest work to ours is \citet{shi2020} which attempted to learn symbolic expression simplification on a dataset of synthetically generated symbolic expressions in Halide by re-writing sub-trees of the parsed expression with reinforcement learning. The domain differs from ours; however, as the domain specific language contains simple expressions and randomly generated programs may contain redundancies not seen in assembly optimized by a compiler like \gcc. 

Another work that addressed automatic program optimization is \citet{bunel2016learning}. Unlike the Halide-based experiments, the work used reinforcement learning to learn a proposal distribution for stochastic search used in \citet{schkufza2013stochastic}. While the learned proposal distribution showed improvements over the baseline, the method ultimately still used stochastic search, except with improved search parameters. Unlike our work, the model is unable to fully control program transformations end-to-end.


\section{Conclusion}

In our work, we explored the task of program superoptimization with neural sequence models. Towards this goal, we utilized 1.61 million programs mined from open source projects on \GH for pre-training along with and a subset of over 25K functions with testcases that can additionally be passed off to the SMT based verifiers in the STOKE project artifacts. We proposed SILO, a learning approach with a two step process (1) an exploration step to search for program superoptimizations, and (2) a learning step on the best sequences found during training. Our experiments on the Big Assembly dataset demonstate that SILO is able to outperform REINFORCE with baseline. We believe that REINFORCE struggles, because program superoptimziation is a highly-challenging exploration task with a very sparse reward space. By incorporating supervision on superoptimized sequences, SILO is able to learn optimizations more effectively from its exploration. 

Recently, large neural sequence models have been proposed as an effective method for program synthesis in high-level programming languages such as Python or C++ from natural language specifications \cite{yin-neubig-2017-syntactic, chen2021evaluating, alphacode}; however, to our knowledge, relatively little work has been done to refine these models to go beyond synthesizing correct programs that meet a specification, and additionally make additional considerations for important metrics such as performance or readability. Given the increased availability of executable program synthesis datasets, tuning neural sequence models to go beyond program synthesis and optimize for additional metrics is a promising direction for future work.  



\bibliography{iclr2022_conference}
\bibliographystyle{iclr2022_conference}

\clearpage

\appendix

\section{Actor-Learner Architecture}

The actor-learner architecture used for training is as follows: before the training process begins, multiple actor threads inheriting the parameters of the parent learner are created. For every iteration, each of the actor threads independently samples a batch of program re-writes from the distribution of the inherited model. 

\begin{figure}[h]
    \centering
    \includegraphics[scale=.60]{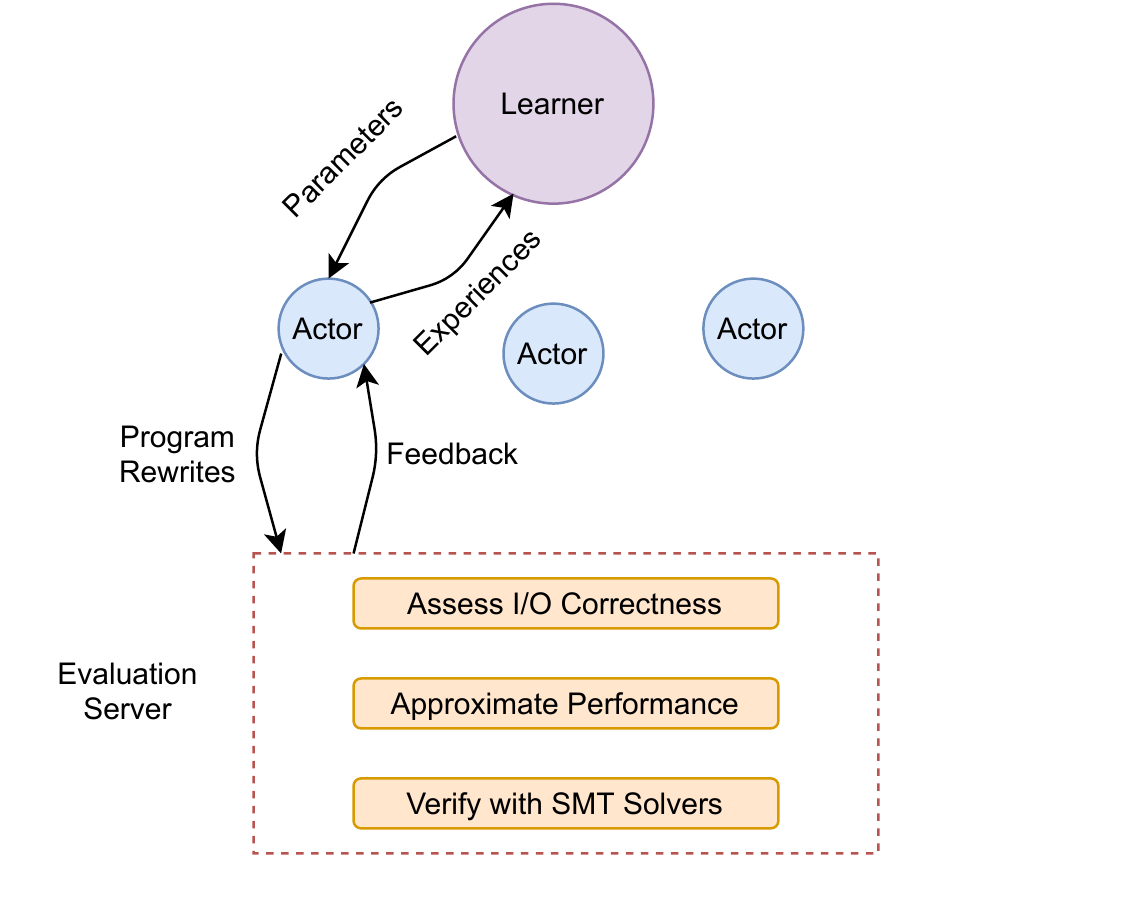}    
    \caption{An overview of our actor-learner setup. It features interaction between a centralized learner module, numerous asynchronously-operating actors, as well as a server for evaluating program rewrites.}
    \label{fig:architecture}
\end{figure}

After sampling a batch of re-writes, an attempt is made the evaluate the rewrites by sending them to an evaluation server. Inside the evaluation server, the programs are assembled and tested for correctness and performance to calculate $\mathcal{C}$ and $\mathcal{V}$; more details on evaluation are located in \cref{sec:evaluation}. The actor then sends the samples with their related cost and correctness information to the learner module for learning. Then, the actor attempts to synchronize its parameters by inheriting a new copy, if available, from the learner module. \cref{fig:architecture} contains an overall diagram for our entire system.

\section{SMT based Verifier Exploits Found}

Before the training process begins, multiple actor threads inheriting the parameters of the parent learner are created. For every iteration, each of the actor threads independently samples a batch of program re-writes from the distribution of the inherited model. 

After sampling a batch of re-writes, an attempt is made the evaluate the rewrites by sending them to an evaluation server. Inside the evaluation server, the programs are assembled and tested for correctness and performance to calculate $\mathcal{C}$ and $\mathcal{V}$; more details on evaluation are located in \cref{sec:evaluation}. The actor then sends the samples with their related cost and correctness information to the learner module for learning. Then, the actor attempts to synchronize its parameters by inheriting a new copy, if available, from the learner module. 

In our manual evaluation stage, we witnessed the primary pattern of exploiting the SMT solver based verifier was that of branch deleation. We present a concrete example of one such exploit paired with the verifier's output in \cref{fig:delete_branch_incorrect}.  In the first subfigure we demonstrate the pattern exploiting the verifier from the original STOKE project artifacts\footnote{\url{https://github.com/StanfordPL/stoke}}, and in the second subfigure we show the output when running on the verifier included in the artifacts from a follow up work on program verification.\footnote{\url{https://github.com/bchurchill/pldi19-equivalence-checker}} In this example,  a comparison is done between the hex constant \texttt{0x2e} and the value located at the address in register \lstinline{

\begin{figure}
    \begin{subfigure}[t]{.6\textwidth}
    \begin{lstlisting}[language={[x64]Assembler}]
Target                   Rewrite              
                                              
.smtp_is_end.s:          .smtp_is_end.s:      
xorl %eax, %eax          xorl %eax, %eax      
cmpb $0x2e, (%rdi)       cmpb $0x2e, (%rdi)   
je .L1                   je .L1               
retq                     .L1:                 
.L1:                     retq                 
movzbl 0x1(%rdi), %edx                        
cmpb $0xd, %dl                                
sete %al                                      
cmpb $0xa, %dl                                
sete %dl                                      
orl %edx, %eax                                
movzbl %al, %eax                              
retq                                          
                                              
Equivalent: yes
\end{lstlisting}
    \caption{Output from the original STOKE bounded verifier}
    \vspace{1em}
    \end{subfigure}
        \begin{subfigure}[t]{.6\textwidth}
    \begin{lstlisting}
Target                   Rewrite              
                                              
.smtp_is_end.s:          .smtp_is_end.s:      
xorl %eax, %eax          xorl %eax, %eax      
cmpb $0x2e, (%rdi)       cmpb $0x2e, (%rdi)   
je .L1                   je .L1               
retq                     .L1:                 
.L1:                     retq                 
movzbl 0x1(%rdi), %edx                        
cmpb $0xd, %dl                                
sete %al                                      
cmpb $0xa, %dl                                
sete %dl                                      
orl %edx, %eax                                
movzbl %al, %eax                              
retq                                          
                                              
[bv] Checking pair: 0 1 2 4; 0 1 2 3
Couldn't take short-circuit option without memory.
[bv] Paths 0 1 2 4 / 0 1 2 3
     verified: true
[bv] Checking pair: 0 1 2 4; 0 1 2 3
Couldn't take short-circuit option without memory.
[bv] Paths 0 1 2 4 / 0 1 2 3
     verified: true
[bv] Checking pair: 0 1 3 4; 0 1 2 3
We've finished early without modeling memory!
[bv] Paths 0 1 3 4 / 0 1 2 3
     verified: true
[bv] Checking pair: 0 1 3 4; 0 1 2 3
We've finished early without modeling memory!
[bv] Paths 0 1 3 4 / 0 1 2 3
     verified: true
Equivalent: yes

\end{lstlisting}
    \caption{Output from verifier in the artifacts from \cite{churchill2019semantic}.} 
    \end{subfigure}
    \centering
    \caption{An example of the common exploit where the right hand side deletes the branch of code following \lstinline{.L1}. If the third and fourth lines (\lstinline{je .L1; .L1:}) are removed, then the verifier actually returns correctly.} 
    \label{fig:delete_branch_incorrect}

\end{figure}

\section{Additional Information on the Big Assembly Dataset}

\paragraph{Data Collection} Our Big Assembly Dataset was mined from open source projects implemented in the C programming language on \GH . Our programs were disassembled using \texttt{GNU} Binutils’ objdump into x86-64 assembly, and split into individual functions. We performed the process twice on the same set of source code, so that we could create a parallel corpus of functions. We split our dataset into train, development, and test sets based at the level of entire individual github projects. We deduplicated by removing any overlapping binaries between the datasets. We also deduplicated at the individual function level by removing string matches after removing function names from assembly functions. Lastly, we removed programs pairs from the dataset if either the source or target program had length greater than 512 after byte-pair encoding tokenization. 

\paragraph{Setting Live out and Filter Spurious Examples} As mentioned in \cref{sec:benchmark} for our each function in our dataset, we were required to determine the \textit{live out} set, the portion of the CPU state required for determining the equivalence between programs. We also determine \textit{heap out}, which is a boolean flag that determines whether or not we should also check the heap for equivalence as well. We perform this sanity check by determining what parts of the CPU state are equivalent between the \gcc \ozero function and the \gcc \othree. Pseudocode for how live out is described in \cref{alg:live_out}, in this algorithm we refer to the \gcc \ozero function as \p. 

In line 1, the algorithm begins by initializing live\_out with all possible CPU registers. In lines 2-5, until either the computation budget is exhausted or the live out set reaches a fixed point, we iteratively execute the function \textit{get\_live\_out} which incrementally determines the candidate live out set. It works by either executing the testcase suite or runs the SMT solver based verifier, analyzing any difference in the resulting CPU state, pruning any part of the CPU state that differs, and then returning the subset of the CPU state that may be equivalent between \p and \pref. This may need to be run iteratively, because after pruning live out with respect to one counter example, it is possible another counterexample may still trigger a difference in other parts of the CPU state. For most of the general purpose CPU registers, we also perform this pruning at the sub-register level, allowing the register size for equivalence to be pruned down to the lower 32, 16, and 8 bits. If the computation budget is exhausted, the program returns early, and the program is discarded from the dataset. 

After determining live out, in lines 6-10 an additional check is done to see if both programs are equivalent when checking the heap. If they are, heap out is set to true, and this information is recorded to be used for the fine-tuning phases. Lastly in lines 12-15, we perform a final sanity check to ensure that none of our programs are equivalent to a set of spurious programs such as a null program, a \texttt{return 0}, and a \texttt{return 1}. If a program is equivalent to one of these highly simplistic programs, it is discarded from our dataset. 

\begin{algorithm}
    \begin{algorithmic}[1]
    \STATE Initialize: $\mathit{live\_out} = \textrm{ALL\_LIVE\_OUT}$ \\
    \REPEAT 
        \STATE $\mathit{old\_live\_out} = \mathit{live\_out}$  \\
        \STATE $\mathit{live\_out} = \mathit{get\_live\_out}(\p, \pref, \mathit{live\_out})$  
      \UNTIL{$\mathit{old\_live\_out} \neq \mathit{live\_out}$ or budget exhausted}  \\
    \IF {$\mathit{diff}(\p, \pref, \mathit{live\_out}, \mathit{heap\_out} = \mathit{True})$}
        \STATE $\mathit{heap\_out} = \mathit{False}$ \\
    \ELSE
        \STATE $\mathit{heap\_out} = \mathit{True}$ \\
    \ENDIF
    \STATE $\mathit{is\_spurious} = \mathit{False}$ \\
    \FOR{ $\mathrm{F}_{\mathit{spur}}$  in $\textrm{SPURIOUS}$} 
        \IF {$ \neg \mathit{diff}(\mathrm{F}_{\mathit{spur}}, \pref, \mathit{live\_out}, \mathit{heap\_out})$}
            \STATE $\mathit{is\_spurious} = \mathit{True} $
        \ENDIF
    \ENDFOR
    \caption{Set Live Out and Filter Examples}
    \label{alg:live_out}
    \end{algorithmic}
\end{algorithm}

\paragraph{Data Preprocessing for Training}

We perform additional processing on our programs that we feed into the model to remove noise; we do this for the \gcc \ozero function $\s$ as well as the \gcc \othree functions used for pre-training $\pref$. x86-64 assembly often uses \textsc{goto}-like instructions to jump to different parts of a binary: this is one way that control flow is implemented. The jump targets are often marked as offsets in the binary itself; however, at the individual function level these may be canonicalized with ordinal locations (i.e. .L1, .L2, and so on). This fully preserves function semantics while removing noise from the prediction task.

\section{Hyperparameters and Settings}

In this section we report the hyperparameters used for fine-tuning our models.

\paragraph{General Hyperparameters}

We found that our SILO algorithm did not need hyperparameter fine-tuning; whereas, our REINFORCE with baseline experiments were more brittle. We witnessed that without a lower learning rate and a carefully tuned learning rate schedule, our REINFORCE experiments would very often diverge before the full fine-tuning budget was exhausted. For all fine-tuning, we used 2,000 steps warmup. For the SILO experiments, we utilized a constant factor of .50 applied to the ``noam scheduler" from \citet{vaswani2017attention}. For the REINFORCE models, we used a factor of .01 for the Big Assembly dataset and a factor of .0025 for the \textsc{Hacker's Delight} dataset. %

\paragraph{REINFORCE Hyperparameters} For our baseline in \cref{eqn:rl_goal}, we used a mean of the objective function for the previous 256 samples for each unique program. After subtracting the mean from the return, we then subsequently normalized by the standard deviation of the objective function of the previous 256 samples. For the lagrangian multiplier $\lambda$ in \cref{eqn:rl_goal}, we used a penalty of 50,000. Additionally, we follow \citet{schkufza2013stochastic} in adding an additional penalty of 100 for every bit in the CPU state that differed between the reference implementations and the rewrite output such that functions with similar semantics would be penalized less than those with  dramatically different semantics.  To prevent our objective function from growing too great, we also clipped the maximum cost so it would not exceed 100,000. As is typical in many policy gradient algorithms, we also included an entropy bonus $\beta$ to encourage additional exploration: we used a constant entropy bonus of $\beta = 0.01$ for both our \textsc{Hacker's Delight} and Big Assembly experiments.

\end{document}